\documentclass[letterpaper, 10 pt, conference]{ieeeconf}  

\IEEEoverridecommandlockouts                              

\usepackage{algpseudocode} 
\usepackage[ruled,vlined]{algorithm2e}
\usepackage{graphicx} 
\usepackage{epsfig}
\usepackage{multirow}
\usepackage{numprint} 
\usepackage{caption}
\usepackage{amsmath,amssymb,amsfonts}
\usepackage[caption=false,font=footnotesize]{subfig}
\usepackage{siunitx}
\usepackage{atbegshi}
\usepackage[absolute,overlay]{textpos}
\usepackage{rotating}
\usepackage[english]{babel}
\usepackage{blindtext}

\usepackage{tikz,xcolor,hyperref}
\definecolor{lime}{HTML}{A6CE39}
\DeclareRobustCommand{\orcidicon}{
	\begin{tikzpicture}
	\draw[lime, fill=lime] (0,0) 
	circle [radius=0.16] 
	node[white] {{\fontfamily{qag}\selectfont \tiny ID}};
	\draw[white, fill=white] (-0.0625,0.095) 
	circle [radius=0.007];
	\end{tikzpicture}
	\hspace{-2mm}
}
\foreach \x in {A, ..., Z}{\expandafter\xdef\csname orcid\x\endcsname{\noexpand\href{https://orcid.org/\csname orcidauthor\x\endcsname}
			{\noexpand\orcidicon}}
}

\title{\LARGE \bf
Reinterpreting Safety Thresholds as Neuron Spiking Thresholds}

\author{Enrico Del Re\orcidE{} \emph{Graduate Student Member, IEEE}, Mohamed Sabry \orcidS{} \emph{Graduate Student Member, IEEE} \\ Cristina Olaverri-Monreal\orcidC{} \emph{Senior Member, IEEE}
\thanks{
        Johannes Kepler University Linz, Austria; Department Intelligent Transport Systems \texttt{\{enrico.del\_re, mohamed.sabry, cristina.olaverri-monreal\}@jku.at}
        }
    }

\begin{document}

\maketitle
\thispagestyle{empty}
\pagestyle{empty}

\begin{abstract}
Surrogate Safety Measures (SSMs) are extensively utilised in the evaluation of traffic risk in automated driving contexts. However, the majority of SSM-based evaluations employ fixed thresholds that fail to capture the human response to sustained borderline conditions or the reaction to brief, high-risk peaks. The present work proposes a biologically inspired reinterpretation of SSM thresholds. This is modelled as spiking thresholds of leaky integrate-and-fire (LIF) neurons, with multiple SSM inputs combined into a spiking neural network (SNN). The SNN is trained to emit spikes that are aligned with human braking onsets. The training data was recorded in a controlled car-following experiment using the 3D-CoAutoSim platform with CARLA/Unreal and a 6-DOF motion platform, where induced critical events were generated. The results demonstrate that the learned spiking activity qualitatively aligns with braking behaviour across scenarios and captures reactions that are not consistently explained by threshold crossings alone. Analysis across participants further indicates that learned input thresholds remain relatively consistent, while learned decay factors encode different temporal sensitivities for the SSMs. The findings of this study indicate that spiking dynamics may serve as a mechanism to facilitate the convergence of objective SSMs with subjective human safety perception.
\end{abstract}


\section{INTRODUCTION}
\label{sec:introduction}

Advances in automated vehicles have largely been driven by the goal of improving traffic safety, as human error contributes to more than 90 $\%$ of all traffic accidents \cite{HumanError}. Advanced Driver Assistance Systems (ADAS), such as adaptive cruise control and lane keeping systems, aim to mitigate these errors and enhance safety \cite{ADAS_impact}. These systems operate according to well-defined objective rules, such as those specified in \cite{UN157}, that connect observable and quantifiable driving parameters (e.g., speed, distance, and relative velocity) through Surrogate Safety Measures (SSMs) to infer accident risk.

However, conventional SSMs exhibit several limitations. Many are tailored to specific interaction types, such as car-following or intersection encounters, and often assume pairwise vehicle interactions, whereas real traffic involves complex multi-agent dynamics. Furthermore, most SSMs rely on categorical thresholding to classify situations as safe or unsafe. Although this approach effectively detects abrupt hazards, such as sudden decelerations, it fails to capture the cumulative strain of borderline conditions sustained over time. These shortcomings become evident when comparing SSM-based safety assessments in naturalistic driving data with human reactions \cite{DSSM,9987124,SSM,10186747}.

Bridging this gap between objective safety indicators and subjective human perception of safety is essential for higher levels of automation, such as platooning, where a human driver may remain in the control loop. Unnecessary takeovers triggered by subjective discomfort are undesirable, and models that ignore human perception risk undermining driver trust and acceptance of ADAS \cite{olaverri2020promoting}. To ensure safe, interpretable, and socially compatible automated behavior, future systems must emulate not only the physical risk but also how humans perceive and react to changing safety conditions.

To address these challenges, this paper proposes a biologically inspired approach that substitutes fixed SSM thresholds with Leaky Integrate-and-Fire (LIF) neurons. These neurons can account for both persistent borderline exposure and sudden high-risk situations. By combining multiple LIF neurons into a Spiking Neural Network (SNN), the framework integrates diverse SSMs into a unified, dynamic safety assessment, akin to a neuron-based extension of \cite{10920258}. The SNN is trained on driving data recorded in an experiment performed with the 3D-CoAutoSim \cite{hussein20183dcoautosim} platform, which utilizes CARLA and Unreal Engine together with a 6-degrees-of-freedom (DOF) motion platform to generate critical traffic scenarios and record realistic human responses. During these scenarios, the braking behavior is recorded, so that, by learning to spike in correspondence with such reactions, the proposed model aligns objective safety metrics with the way humans subjectively perceive and respond to traffic risk.

The following section provides an overview of comparable approaches, emphasizing the unique approach in our proposed model. Section \ref{sec:Background} introduces the LIF neuron, SNNs and the utilized SSMs. Section \ref {sec:Methodology} details the experiment and training details. The results are presented in Section \ref{sec:Results}. Lastly, Section \ref{sec:Conclusion} summarizes the key findings and outlines potential future directions for this research.

\section{Related Work}
\label{sec:RelatedWork}

SNNs in automotive and transportation have been mostly focused on perception tasks, to take advantage of their inherent energy efficiency and high inference speed \cite{SNN_review}. Less research has focused on predicting human behavior, though for example \cite{SNN_pedestrian} used SNNs to predict pedestrian street-crossing behavior and \cite{SNN_medical} used SNNs to predict turn-taking in surgeries.

Modeling the subjective perception of safety presents a notable challenge due to a scarcity of labeled data, specifically, data encompassing both objective and subjective measures.  In \cite{SSM_model_india} such a dataset was created by conducting assessments of traffic conflicts by trained observers. This method enabled the identification of objective factors predominantly influencing individuals' safety perceptions, specifically those affecting both the proximity and severity of potential collisions. However, the labor-intensive nature of labeling interactions makes this approach less practical for more complex scenarios.

Perceived safety has also been addressed by proposing risk representations intended to better match what humans internally evaluate, and validating them against observable human reactions, most commonly in simulations. Risk-field formulations model risk as a spatial field shaped by predicted motions and interaction consequences, and have been reported to correlate with subjective markers such as verbalized risk. The steering behavior observed in \cite{experiment_for_Chen} has been used by \cite{Chen_2024_POLAR} to align a model to individual participants’ steering behavior. In \cite{kolekar2020human} a driver’s risk field metric was proposed to model human driving behavior.

However, risk-field-based approaches often do not map cleanly back to standard SSMs, such as Time Headway or Time to Collision, which limits their direct use within established AV safety evaluation guidelines. Therefore, the model presented in this paper explicitly leverages common surrogate safety metrics as inputs for modeling driving behavior, letting the biologically inspired network learn the transformation toward human behavior over time, rather than redefining a new risk metric upfront.

\section{Background}
\label{sec:Background}

\subsection{Spiking Neural Networks}

\begin{figure}
    \centering
    \includegraphics[width=0.9\linewidth]{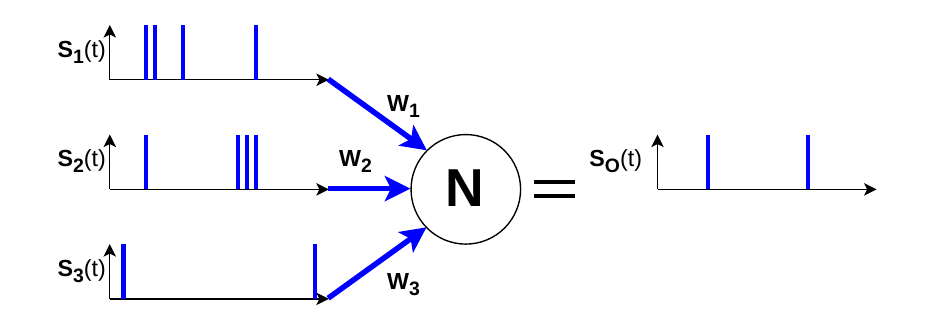}
    \caption{Schematic propagation of spikes through a neuron in an SNN. The three spike trains $S_{1/2/3}(t)$ from a previous layer are multiplied by their respective weights $W_{1/2/3}$ and processed by the neuron N into an new spike train $S_{O}(t)$.}
    \label{fig:SNN_schematic}
\end{figure}

Spiking Neural Networks (SNNs) process information as time-indexed spike trains rather than continuous activations. In contrast to standard artificial neural networks, computation unfolds over discrete timesteps and each neuron maintains an internal state (typically a membrane potential) that integrates inputs over time. Once a potential threshold is exceeded, a spike by the neuron is emitted. In a feed-forward SNN, these spikes are weighted by synapses and integrated by spiking neuron dynamics, producing new spike trains that propagate through the layers (Fig.~\ref{fig:SNN_schematic}). This temporal state makes SNNs well-suited for sequence data, as they can naturally capture accumulation effects and short-lived peaks via their dynamics.

\subsubsection{Leaky-integrate-and-fire (LIF) neuron}
\begin{figure}
    \centering
    \includegraphics[width=0.8\linewidth]{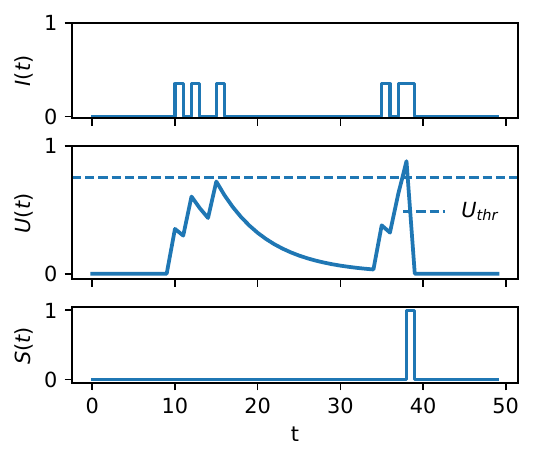}
    \caption{Exemplary behavior of a LIF-neuron. The input current $I(t)$ is integrated by the neuron as membrane potential $U(t)$, which results in the spike train $S(t)$. Note how the membrane potential decays over time slowly, yet resets to 0 immediately after a spike, when $U(t)$ has crossed the spiking threshold $U_{thr}$.}
    \label{fig:LIF}
\end{figure}

Although many neuron models are suitable for SNNs, here we limit ourselves to leaky-integrate-and-fire (LIF) neurons \cite{LIF_review}. In discrete time, the membrane potential $U_t$
integrates an input current $I_t$ while decaying with factor $\beta$:
\begin{equation}
U_t = \beta U_{t-1} + I_t.
\label{eq:lif_discrete}
\end{equation}
Here, $\beta$ controls the effective memory of the neuron: values closer to 1 yield slower decay and longer temporal integration, while smaller values emphasize more immediate inputs. A spike is emitted when $U_t$ crosses a threshold $U_{\mathrm{thr}}$:
\begin{equation}
S_t =
\begin{cases}
1 & \text{if } U_t \ge U_{\mathrm{thr}}, \\
0 & \text{otherwise.}
\end{cases}
\label{eq:heaviside_spike}
\end{equation}
After spiking, the membrane potential is reset, most commonly to 0. The interplay between input current, membrane potential and output spike train is visualized in Figure \ref{fig:LIF}

\subsubsection{Training with surrogate gradients}
Because Eq.~\eqref{eq:heaviside_spike} is non-differentiable, training is commonly done using backpropagation-through-time (BPTT) \cite{BPTT} with a
surrogate gradient. The network is unrolled over timesteps and gradients are accumulated across time similarly to recurrent networks. In the forward pass the hard threshold is kept (Eq.~\eqref{eq:heaviside_spike}), while in the backward pass $\partial S_t/\partial U_t$ is approximated with a surrogate, e.g. a fast-sigmoid :
\begin{equation}
\frac{\partial S}{\partial U} \approx \frac{1}{\bigl(1 + k \lvert U \rvert \bigr)^{2}},
\qquad k=25.
\label{eq:surrogate}
\end{equation}
This approximation enables gradient-based optimization while preserving event-driven spiking in the forward dynamics.

\subsection{Surrogate Safety Measures}

To assess safety between two vehicles in a comprehensive way, in scenarios involving trajectory intersections and car-following situations, we incorporated the following metrics into our safety model.
\begin{itemize}
\item Time Headway (TH): Duration for the rear of one vehicle to traverse a specific point on the road, followed by the front of the subsequent vehicle passing the same reference point.
\item Time-To-Collision (TTC): Time for a vehicle to reach a potential collision point if it maintains its current speed and trajectory. 
\item Minimum Deceleration Required to Avoid Crash (DRAC): Lowest deceleration rate that a vehicle must achieve in order to prevent a collision with another vehicle.

\end{itemize}

All three SSMs were selected for their suitability in car-following scenarios. Since the SNN is trained to spike under high-risk conditions, the inputs are transformed so that larger input values correspond to objectively higher safety risk. For TH we take the inverse according to \eqref{eq2},

\begin{equation}
\label{eq2}
    \frac{1}{TH} = \frac{v}{d},
\end{equation}
with $v$ the velocity of the following vehicle and $d$ the distance to the leading vehicle.

The inverse for TTC as been introduced as an SSM in ~\cite{ITTC}, as denoted in equation~\eqref{eq3},

\begin{equation}
\label{eq3}
    ITTC = \frac{v_F-v_L}{d},
\end{equation}

where $v_{L}, v_F$ represents the velocity of the leading and following vehicles and $d$ denotes the distance between them.  
If the two vehicles are not on a collision course, ITTC is set to $0$.\\

DRAC is is computed as defined in ~\eqref{eq4},

\begin{equation}
\label{eq4}
    DRAC = \begin{cases}
    \frac{(v_F-v_L)^2}{d}, & v_F > v_L \\
    0, & else
    \end{cases}
\end{equation}.

\section{Methodology}
\label{sec:Methodology}

\subsection{Dataset}

The aforementioned lack of driving-behavior datasets annotated with subjective safety perception limits learning-based approaches. While training on subsets of public datasets, e.g. HighD \cite{highD} or IAMCV \cite{icmcv} is desired, these datasets make it difficult to attribute human braking behavior primarily to safety concerns. Therefore, a new dataset was recorded in which participant interventions, limited to braking, were largely restricted to controlled changes in traffic risk. Part of this dataset is used to train the model in this paper.

\begin{figure}
    \centering
    \includegraphics[width=0.9\linewidth]{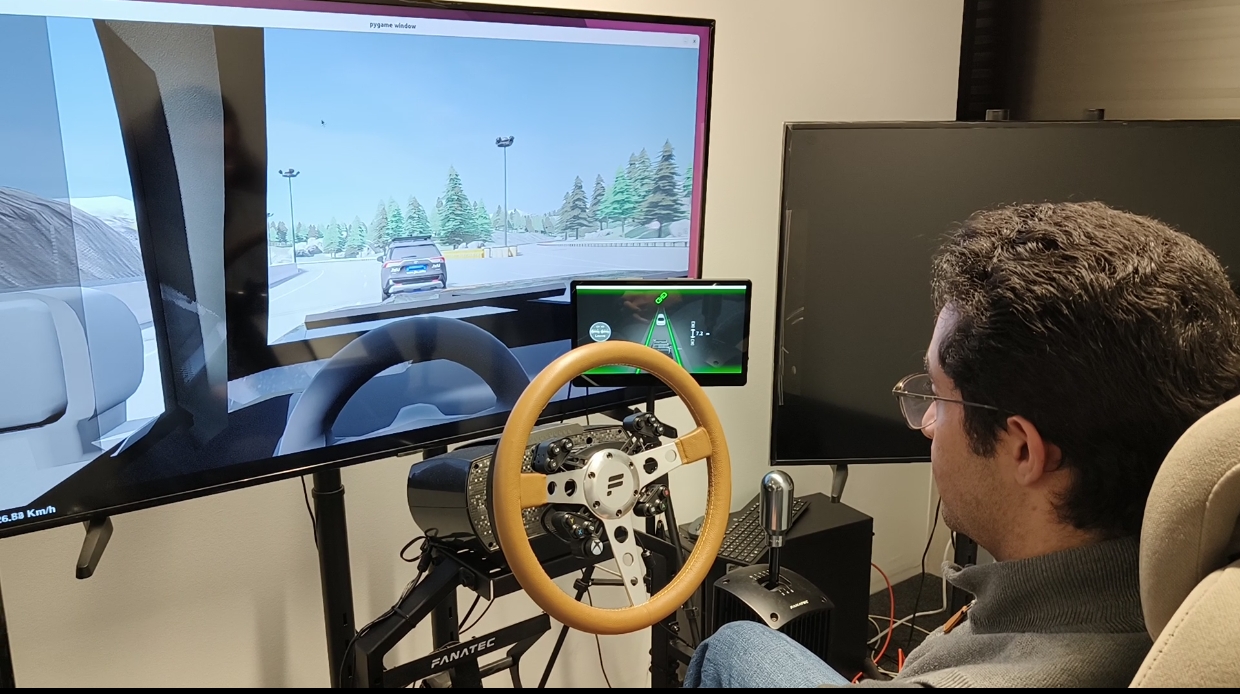}
    \caption{The 3DCoAutoSim Simulation setup with CARLA.}
    \label{fig:simulator}
\end{figure}

The simulation environment consisted of a 6-degrees-of-freedom motion platform connected to the CARLA simulator \cite{CARLA} as shown in Figure \ref{fig:simulator}. Participants experienced a car-following scenario on a highway in the follower vehicle in which, after reaching the cruising speed of 80 kph, the autonomous leading vehicle would begin to decelerate at a random point within a 70 m range window, increasing the safety risk for the following vehicle. To ensure consistent initial conditions before the lead-vehicle deceleration, the participant's vehicle was initially controlled using the controller from \cite{sabry_sim}. The CARLA simulation developed for validation in \cite{sabry_sim} was also used as the baseline for the scenarios in this simulation.

Participants could intervene at any time by pressing the brake if they felt unsafe, steering was only possible in conjunction with the brake. After the leading vehicle started braking, the controller still attempted to maintain the desired distance, though a subset of trials was intentionally configured to result in a collision unless the participant intervened. The motion platform further enhanced the realism of the deceleration cues and the perceived urgency of these situations. 

The six scenarios were defined by the deceleration of the leading vehicle and the controller's desired following distance:
\begin{itemize}
    \item \textbf{Scenario 1:} Deceleration: $-6m/s^2$, distance: $10 m$
    \item \textbf{Scenario 2:} Deceleration: $-6m/s^2$, distance: $5 m$
    \item \textbf{Scenario 3:} Deceleration: $-3m/s^2$, distance: $10 m$
    \item \textbf{Scenario 4:} Deceleration: $-3m/s^2$, distance: $5 m$
    \item \textbf{Scenario 5:} Deceleration: $-4.5m/s^2$, distance: $10 m$
    \item \textbf{Scenario 6:} Deceleration: $-4.5m/s^2$, distance: $5 m$
\end{itemize}

The order of scenarios was randomized but identical across participants. At the start of the experiment, each participant was given time to familiarize themselves with the simulator and the vehicle dynamics in a different driving scenario. In total, the recordings of 9 participants were utilized for training.Each scenario took around 40 s to complete, depending on driver intervention.

Based on these recordings, the data was segmented into episodes (i.e., per participant and scenario). Each episode forms a multivariate time series $\mathbf{x}_{1:T}\in\mathbb{R}^{T\times 5}$ with the three input features 1/TH, ITTC and DRAC, the brake value of the participant and timestamps. 
Given the limited number of episodes per participant, training is treated as a per-participant optimization procedure rather than a classical train/val/test benchmark.

\subsection{Spiking Neural Network (SNN) Architecture and Implementation Details}
We implemented the model in Python using PyTorch and the \texttt{snntorch} library, which provides standard leaky integrate-and-fire (LIF) neuron dynamics and surrogate-gradient training \cite{snntorch}. At each time step $t$, the network receives a 3-dimensional input vector $\mathbf{x}_t \in \mathbb{R}^{3}$ consisting of inverse time headway ($1/\mathrm{TH}$), ITTC, and DRAC. The input sequence $\{\mathbf{x}_t\}_{t=1}^{T}$ is encoded by three input neurons with an initial threshold of 1, 1/1.5, 3.3 respectively, based on traffic safety thresholds \cite{DSSM}, and processed by a fully connected feed-forward SNN with two hidden layers followed by a single output neuron.

Each layer applies a linear transformation followed by the LIF dynamics described in Section~\ref{sec:Background}, using trainable neuron thresholds and trainable decay factors $\beta$ for all neurons, bias terms are omitted. The network is trained with BPTT using surrogate gradients (fast-sigmoid surrogate as in \ref{eq:surrogate}).

Training is performed per participant, as in \cite{Chen_2024_POLAR}, by grouping the episodes of each participatn and zero-padding all sequences to the maximum episode length $T$ of the respective participant for batch processing. Adam \cite{ADAM} is used for optimization, using a learning rate selected via a predefined sweep, and train for up to 1000 epochs. An evaluation pass is performed each epoch and the checkpoint with the lowest validation loss is retained.

Weights of all linear layers are initialized randomly from $\mathcal{U}(0,1)$ and constrained to be non-negative to support an "excitatory-only" interpretation of connections.
A ReduceLROnPlateau scheduler monitors the validation loss and reduces the learning rate by a factor of $0.1$ after 5 epochs without improvement. Early stopping is applied if the validation loss does not improve by at least $10^{-6}$ for 20 consecutive epochs.

Model selection is performed via a grid search over hidden size $H\in\{8,16\}$ and learning rate $\eta\in\{10^{-2},10^{-3},5\cdot10^{-4}\}$ (up to 1000 epochs).

\subsection{Target and Loss Function}
The supervision signal is derived from the participant's brake pedal signal  and is intended to represent the intensity and temporal footprint of braking onsets.
Let $b_t \in [0,1]$ denote the brake signal at timestamp $t$.
We compute the discrete rate of change (a time-derivative) as
\begin{equation}
r_t = \frac{b_t - b_{t-1}}{\Delta t_t},
\end{equation}
where $\Delta t_t$ is the time step between successive timestamps.
We focus on increases in brake activation by thresholding $r_t$:
\begin{equation}
m_t = 
\begin{cases}
A_{\mathrm{fac}}\, r_t, & \text{if } r_t \ge r_{\mathrm{thr}},\\
0, & \text{otherwise,}
\end{cases}
\end{equation}
where $r_{\mathrm{thr}}$ is a threshold to avoid random noise and $A_{\mathrm{fac}}$ is a scaling factor.

Although BPTT is used for training, due to convergence stability and speed, hebbian learning and spike-timing-dependent plasticity (STDP) \cite{Hebbian_2008} rules were a natural first consideration since the goal of the network is to detect and ideally predict human braking onset. We therefore define an exponential envelope that serves as a decaying event trace, analogous to eligibility traces used in biologically inspired plasticity models, while training the network end-to-end with supervised gradient descent.

The exponential envelope spreads each event backward in time:
\begin{equation}
y_t = \sum_{k=t}^{T} m_k \exp\!\left(-\frac{k-t}{\tau}\right),
\end{equation}
where $\tau$ is a decay constant.
Intuitively, larger positive brake increases produce a larger envelope amplitude, and the exponential term assigns higher values closer to the braking onset.

With $\hat{y}_t$ denoting the network output at time $t$ (the output neuron spike signal, treated as a time series), we optimize the model using the mean squared error (MSE) loss function over the full sequence:
\begin{equation}
\mathcal{L}_{\mathrm{MSE}} = \frac{1}{T}\sum_{t=1}^{T}\left(\hat{y}_t - y_t\right)^2.
\end{equation}
This objective encourages the network to align its spiking activity with the braking-onset envelope derived from the participant's brake dynamics.

\section{Results}

\label{sec:Results}

\subsection{Alignment with Braking Onsets}
\begin{figure*}[t]
    \centering

    \subfloat[Scenario 1]{%
        \includegraphics[width=0.32\textwidth]{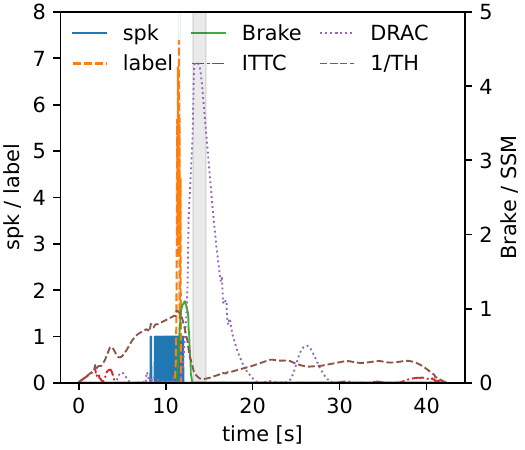}
        \label{fig:a}
    }\hfill
    \subfloat[Scenario 2]{%
        \includegraphics[width=0.32\textwidth]{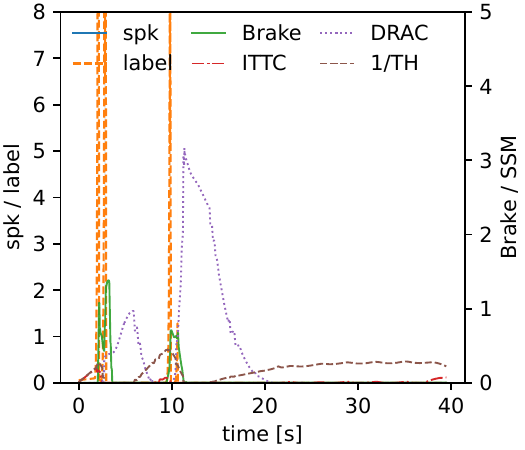}
        \label{fig:b}
    }\hfill
    \subfloat[Scenario 3]{%
        \includegraphics[width=0.32\textwidth]{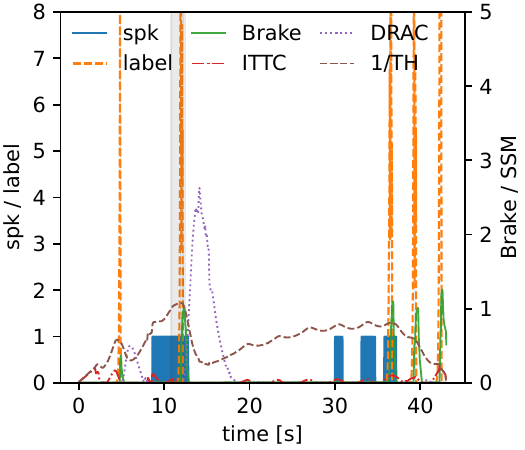}
        \label{fig:c}
    }\\

    \vspace{-1mm} 

    \subfloat[Scenario 4]{%
        \includegraphics[width=0.32\textwidth]{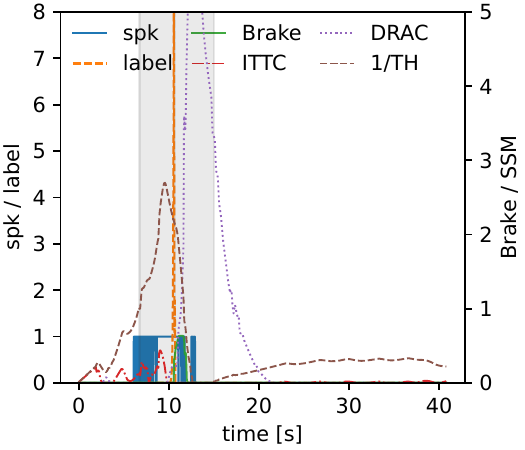}
        \label{fig:d}
    }\hfill
    \subfloat[Scenario 5]{%
        \includegraphics[width=0.32\textwidth]{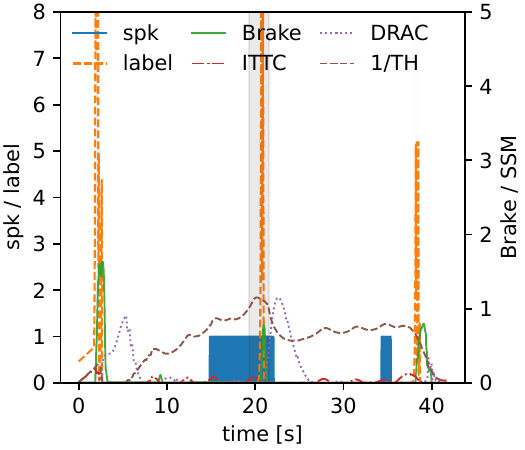}
        \label{fig:e}
    }\hfill
    \subfloat[Scenario 6]{%
        \includegraphics[width=0.32\textwidth]{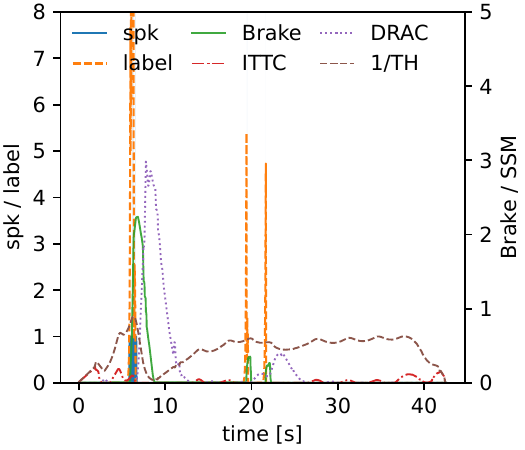}
        \label{fig:f}
    }

    \caption{Objective safety measures (1/TH, ITTC, DRAC) and braking for single participant, along the targets (labels) and predictions (spk) by the trained SNN model for Scenario 1-6. In grey is marked the area where SSM thresholds mark the traffic scenario as unsafe.}
    \label{fig:part_scenarios}
\end{figure*}

Figure \ref{fig:part_scenarios} shows the performance of a trained model for one participant across all six scenarios. While the model occasionally produces spiking activity in periods without braking (e.g., Figure \ref{fig:c}) and sometimes fails to detect some braking onsets (e.g., Figure \ref{fig:b}), it generally aligns well with the overall timing and structure of the participant’s braking behavior. The figure also contrasts the model output with purely threshold-based safety assessments. Regions deemed unsafe by TH, ITTC, and DRAC are highlighted in grey, with darker shading indicating that multiple safety thresholds are exceeded simultaneously. Notably, Figures \ref{fig:b}, \ref{fig:c}, \ref{fig:e}, and \ref{fig:f} show cases where SSM thresholds and human reactions differ, yet the model captures several braking responses that are not directly explained by threshold crossings alone.

This configuration used two hidden layers with eight neurons each and was trained with an exponential envelope defined by $A_{\mathrm{fac}}=1.0$ and $\tau=2.0$, using a learning rate of $10^{-2}$.

\subsection{SSM Spiking Thresholds}

The best-performing model for each participant was selected for analysis of overarching effects across all participants. Since $A_{frac}$ and $\tau$ were also considered in the parameter search, the best model was selected based MSE and on the fraction of output spikes during the braking maneuver to mitigate differences in the loss envelope.

\begin{figure}
    \centering
    \includegraphics[width=0.95\linewidth]{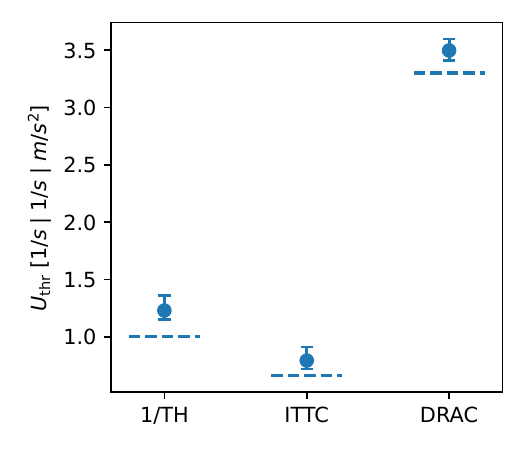}
    \caption{Mean spiking thresholds of the input layer LIF-neurons for the three input values. The error bars mark the lowest and highest value across all participants. The dashed line mark the initial values from literature.}
    \label{fig:thresholds}
\end{figure}

\begin{figure}
    \centering
    \includegraphics[width=0.95\linewidth]{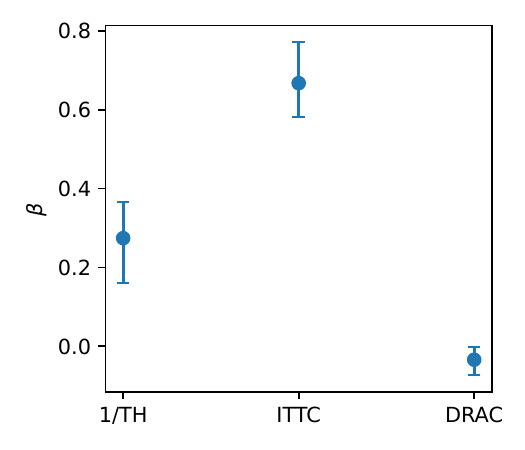}
    \caption{Mean membrane decay constant of the input layer LIF-neurons for the three input values. The error bars mark the lowest and highest value across all participants.}
    \label{fig:beta}
\end{figure}

The initial thresholds were set based on values from literature. Figures~\ref{fig:thresholds} and \ref{fig:beta} show the mean, minimum, and maximum of the trained thresholds of the input LIF neurons and their decay rates, respectively. Across all SSM inputs, the learned thresholds increase slightly relative to their initialization and remain broadly consistent across participants. In terms of temporal dynamics, non-zero ITTC values are retained for the longest period of time (highest $\beta$), whereas the related collision-severity proxy DRAC is forgotten almost immediately (lowest $\beta$). Inverse time headway (1/TH) exhibits faster decay than ITTC as well; however, since 1/TH remains non-zero during car-following, a shorter effective memory is expected.

\section{Conclusion and Future Work}
\label{sec:Conclusion}

This paper proposes reinterpreting fixed thresholds for surrogate safety measures as trainable spiking thresholds within a leaky integrate-and-fire (LIF) framework. By integrating multiple SSM streams with an SNN and training it to spike in correspondence with human braking onsets, the approach translates objective safety indicators to subjective human reactions. Training was performed using driving data recorded in a controlled car-following experiment on the 3D-CoAutoSim platform utilizing CARLA and a 6 DOF motion platform. We demonstrated that the trained models could align with the braking behavior of participants across multiple lead-vehicle deceleration scenarios and, in particular, could reproduce braking responses in situations where simple SSM thresholds were insufficient.

However, the low number of both scenarios and participants did not allow generalization and limited the training to an optimization problem without generalization. In addition, safety perception in simulations can differ substantially to real-world scenarios. Similarily, the limited sample size was insufficient for generalizations.
Therefore, future work will extend training beyond per-participant models and will evaluate generalization to richer multi-agent scenarios and additional SSM inputs. In particular, validation on real-world driving data is planned to further assess the applicability of the proposed approach.



\bibliographystyle{IEEEtran}
\bibliography{bliblio}

\end{document}